\documentclass[NewProceedings,letterpaper]{ascelike-new}
\usepackage[utf8]{inputenc}
\usepackage[T1]{fontenc}
\usepackage{lmodern}
\usepackage{graphicx}
\usepackage[figurename=Figure,labelfont=bf,labelsep=period]{caption}
\usepackage{subcaption}
\usepackage{amsmath}
\usepackage[document]{ragged2e}
\usepackage{newtxtext,newtxmath}
\usepackage[dvipsnames]{xcolor}
\usepackage[colorlinks=true,citecolor=black,linkcolor=black,urlcolor=RawSienna]{hyperref}

\begin{document}

\title{Read the Room: Inferring Social Context Through Dyadic Interaction Recognition in Cyber-physical-social Infrastructure Systems}
    
\author[1]{Cheyu Lin}
\author[2]{John Martins}
\author[3]{Katherine A. Flanigan, Ph.D.}

\affil[1]{\raggedright{Department of Civil and Environmental Engineering, Carnegie Mellon University, 5000 Forbes Ave, Pittsburgh, PA 15213; e-mail: \href{cheyul@andrew.cmu.edu}{cheyul@andrew.cmu.edu}}}
\affil[2]{Department of Civil and Environmental Engineering, Carnegie Mellon University, 5000 Forbes Ave, Pittsburgh, PA 15213; e-mail: \href{johnmart@andrew.cmu.edu}{johnmart@andrew.cmu.edu}}
\affil[3]{Department of Civil and Environmental Engineering, Carnegie Mellon University, 5000 Forbes Ave, Pittsburgh, PA 15213; e-mail: \href{kflaniga@andrew.cmu.edu}{kflaniga@andrew.cmu.edu} (Corresponding author)}

\maketitle

\justifying
\vspace{-0.5cm}
\begin{abstract}
\vspace{\baselineskip}
\noindent
Cyber-physical systems (CPS) integrate sensing, computing, and control to improve infrastructure performance, focusing on economic goals like performance and safety. However, they often neglect potential human-centered (or ``social'') benefits. Cyber-physical-social infrastructure systems (CPSIS) aim to address this by aligning CPS with social objectives. This involves defining social benefits, understanding human interactions with each other and infrastructure, developing privacy-preserving measurement methods, modeling these interactions for prediction, linking them to social benefits, and actuating the physical environment to foster positive social outcomes. This paper delves into recognizing dyadic human interactions using real-world data, which is the backbone to measuring social behavior. This lays a foundation to address the need to enhance understanding of the deeper meanings and mutual responses inherent in human interactions. While RGB cameras are informative for interaction recognition, privacy concerns arise. Depth sensors offer a privacy-conscious alternative by analyzing skeletal movements. This study compares five skeleton-based interaction recognition algorithms on a dataset of 12 dyadic interactions. Unlike single-person datasets, these interactions, categorized into communication types like emblems and affect displays, offer insights into the cultural and emotional aspects of human interactions.
\end{abstract}
\vspace{4pt}
\section{Introduction} 
\vspace{\baselineskip}
\noindent Cyber-physical systems (CPS) have been revolutionizing infrastructure management for years, combining sensing, computing, and control to enhance system performance. While these systems are successful at meeting economic goals such as efficiency and safety, they overlook an important aspect: the human-centered (or ``social'') benefits that arise from infrastructure \cite{annaswamy_control_nodate}. Public open spaces, for example, are crucial components of infrastructure that play a vital role in enhancing community life. They serve as recreational areas where people engage in various activities \cite{koohsari_public_2015}, fostering social connections and a sense of community belonging \cite{kazmierczak_contribution_2013}. Well-designed public spaces improve mental health, physical health, and overall well-being \cite{wang_assessing_2021}. 
Although countless studies link social benefits to the design and condition of infrastructure \cite{salih_criteria_2017}, there is a lack of concrete, measurable knowledge on how to effectively incorporate and enhance social systems within CPS.

\begin{figure} [t]
    \begin{center}
        \begin{tabular}{c}
             \includegraphics[height=3.6cm, width=10.8cm]{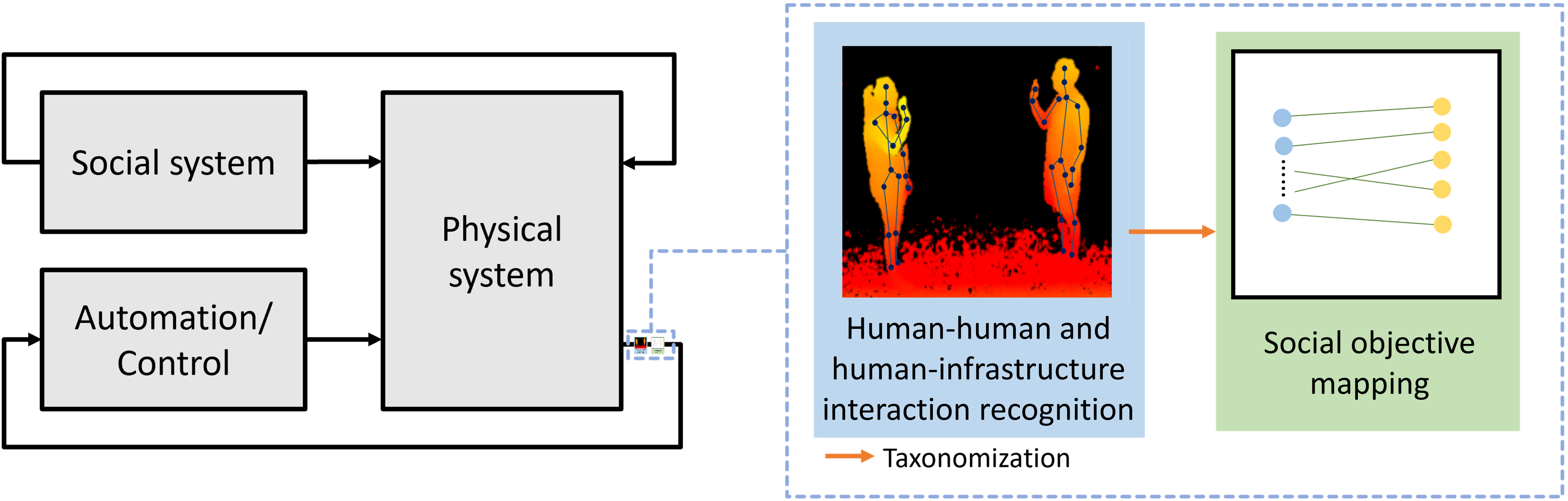} 
        \end{tabular}
    \end{center}
    \vspace{-0.75cm}\caption{\textbf{Situation of work within CPSIS \protect\cite{annaswamy_control_nodate}.}}\vspace{-0.7cm}
    \label{fig:CPSIS}
\end{figure}

\citeN{doctorarastoo_exploring_2023} terms such an infrastructure system that is designed and controlled to meet social objectives as a cyber-physical-social infrastructure system (CPSIS) (Figure \ref{fig:CPSIS}). To develop a CPSIS, it is necessary to define social benefits derived from the built environment, classify human-human and human-infrastructure interactions, develop responsible (i.e., privacy preserving) means of measuring these interactions, model interactions for prediction, map interactions to social benefits, and actuate the physical environment to promote desirable social outcomes \cite{doctorarastoo_exploring_2023}. Hence, the research challenges needed to realize this paradigm shift can be broadly grouped into four categories. (1) Behavioral modeling: Models based on behavioral and cognitive principles focused on understanding and representing the connections between characteristics of the built environment and the behaviors or actions that result from these characteristics. (2) Human dynamics modeling: Predictive models that show how people react to different stimuli in their physical surroundings. (3) Indirect social monitoring: This area aims to classify social activities---as well as human-human and human-infrastructure interactions---and understand deeper meanings inherent in these interactions. (4) Control theory: Control of highly complex, dynamic systems characterizing social response to infrastructure adaptation.

The underlying scaffolding for solutions to the four research challenges is the ability to identify and measure human-human and human-infrastructure interactions from real-world, \textit{in situ} data. There are already many existing modalities for human interaction recognition (HIR), such as RGB cameras and depth sensors \cite{motiian_online_2017}. RGB cameras have been used ubiquitously for HIR. 
While cameras provide an abundance of information, they give rise to the unauthorized use of personal information \cite{barhm_negotiating_2011}. Depth sensors have ushered in a new area of research focusing on recognizing interactions through skeletal movements while prioritizing the protection of privacy. The use of 3D key-point coordinates for skeletons not only improves computational efficiency, but also adds a third dimension that helps to overcome the issue of varying camera angles, a common limitation of camera-based systems. However, little is known about how computer vision algorithms are best translated to depth sensor analysis, especially given the specific processing goals and contexts.  Comparing the performance of different algorithms and datasets helps facilitate the advancement of skeleton-based approaches. Benchmarking has already helped assess and uncover the different capabilities of skeleton-based algorithms for single-person---or \textit{monadic}---activities, justifying the use of specific models \cite{martins_skeleton-based_2023}. The MPII dataset \cite{andriluka_2d_2014}, for example, includes 410 videos of random actions (i.e., daily activities) from YouTube. However, a similar holistic dataset has yet to be presented for the study of \textit{dyadic} interactions. The spatial and temporal coordination that reveals interaction context increases interaction variation greatly \cite{stergiou_analyzing_2019}, making it impossible for one dataset to exhaust all possible interactions. There is a need to enhance understanding of the deeper meanings and mutual responses inherent in human interactions, offering a more comprehensive view of social dynamics.

To systematically enable the indirect monitoring of complex, dyadic human interactions in a privacy-preserving way, there is a need to first develop a dataset that contextually categorizes human interactions and quantifies the performance of different algorithms for interpreting the dataset. To meet this need, we benchmark 5 skeleton-based interaction recognition algorithms on a dataset that comprises 12 human interactions collected at Carnegie Mellon University (CMU). The 5 algorithms include convolutional neural network (CNN), bi-directional long short-term memory network (Bi-dir LSTM), convolutional LSTM (ConvLSTM), spatio-temporal graph convolutional network (ST-GCN), and transformer. Unlike existing datasets that include arbitrary daily activities of a single person, these 12 dyadic interactions are selected from a classification system \cite{ekman_repertoire_2010}, identifying 5 types of human interactions that have communication functions: emblems, illustrators, affect displays, regulators, and adaptors. This methodological choice allows for a more nuanced understanding of human interactions, not just in their physical manifestation but also in their cultural, emotional, and interactive significance.  By focusing on specific, taxonomized interaction types, this research moves beyond the mere tracking of body movements, contributing to a deeper comprehension of the embedded meanings and reciprocities in human interactions. This work sets the stage for developing more sophisticated and context-aware interaction recognition systems for the study of CPSIS. 


\begin{figure} [b]
    \begin{center}
        \begin{tabular}{c c c c}
             \includegraphics[height=2.24cm, width=2.8cm]{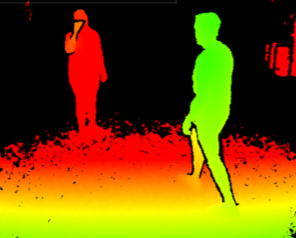} & 
             \includegraphics[height=2.24cm, width=2.8cm]{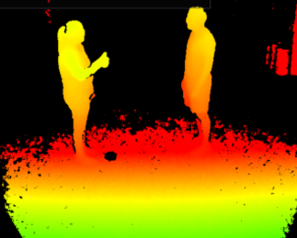}  &
             \includegraphics[height=2.24cm, width=2.8cm]{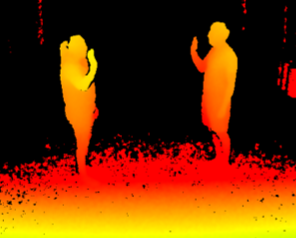}  &
             \includegraphics[height=2.24cm, width=2.8cm]{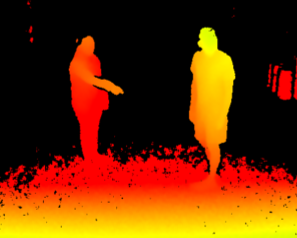}  \\
             \includegraphics[height=2.24cm, width=2.8cm]{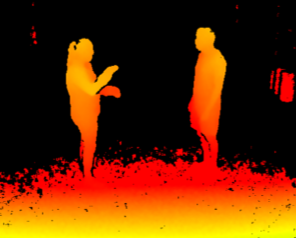}  &
             \includegraphics[height=2.24cm, width=2.8cm]{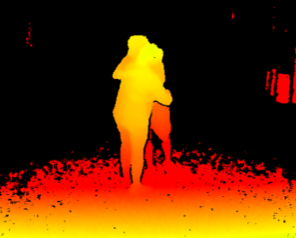}  &
             \includegraphics[height=2.24cm, width=2.8cm]{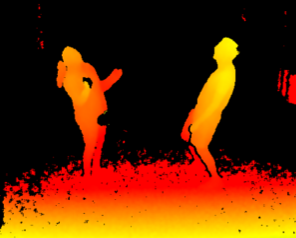}  &
             \includegraphics[height=2.24cm, width=2.8cm]{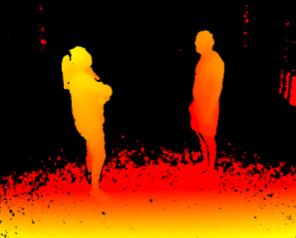}  \\
             \includegraphics[height=2.24cm, width=2.8cm]{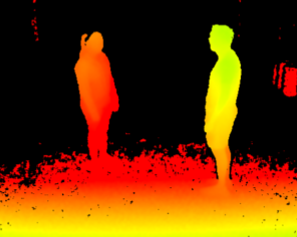}  &
             \includegraphics[height=2.24cm, width=2.8cm]{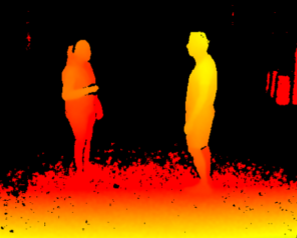}  &
             \includegraphics[height=2.24cm, width=2.8cm]{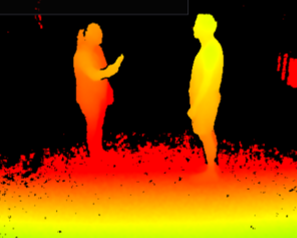}  &
             \includegraphics[height=2.24cm, width=2.8cm]{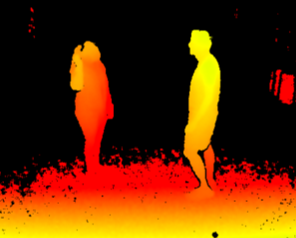}
        \end{tabular}
    \end{center}
    \vspace{-0.7cm} \caption{\justifying \textbf{Sample depth data of 12 interactions. From left to right and top to bottom: Waving in, thumbs-up, waving, pointing, showing measurements, hugging, laughing, arm crossing, nodding, writing circles in the air, holding palms out, and twirling or scratching hair.}} \vspace{-0.7cm}
    \label{fig:depth_data}
\end{figure}
\section{Contextualizing Human Interactions}
\vspace{\baselineskip}
\noindent Interaction is defined as a dynamic sequence of social actions between two or more individuals that can implicitly exchange information. We recognize that the integration of human-human and human-infrastructure interaction data into a CPSIS framework that can map interactions to social benefits requires the study of specific, categorized types of interactions; the decision of which interactions to study is not arbitrary. With such a classification system, we can taxonomize and study interactions more systematically instead of attempting to exhaust all human interactions. There are many channels for reciprocation, such as verbal communication, facial expression, and bodily movements \cite{lopez_postural_2017}.  In the context of CPSIS, bodily movements are the most suitable option to study since they can be represented by anonymized skeletons \cite{cowan_san_2019}. 

A total of 12 interactions are adopted from the classification system developed by \citeN{ekman_repertoire_2010} as shown in Figure \ref{fig:depth_data}. The system arranges human interactions into five categories based on their functional interaction movements, which include
emblems, illustrators, affect displays, regulators, and adaptors.  

\begin{figure} [b]
    \begin{center}
        \begin{tabular}{c c c}
             \includegraphics[height=5cm, width=3.5cm]{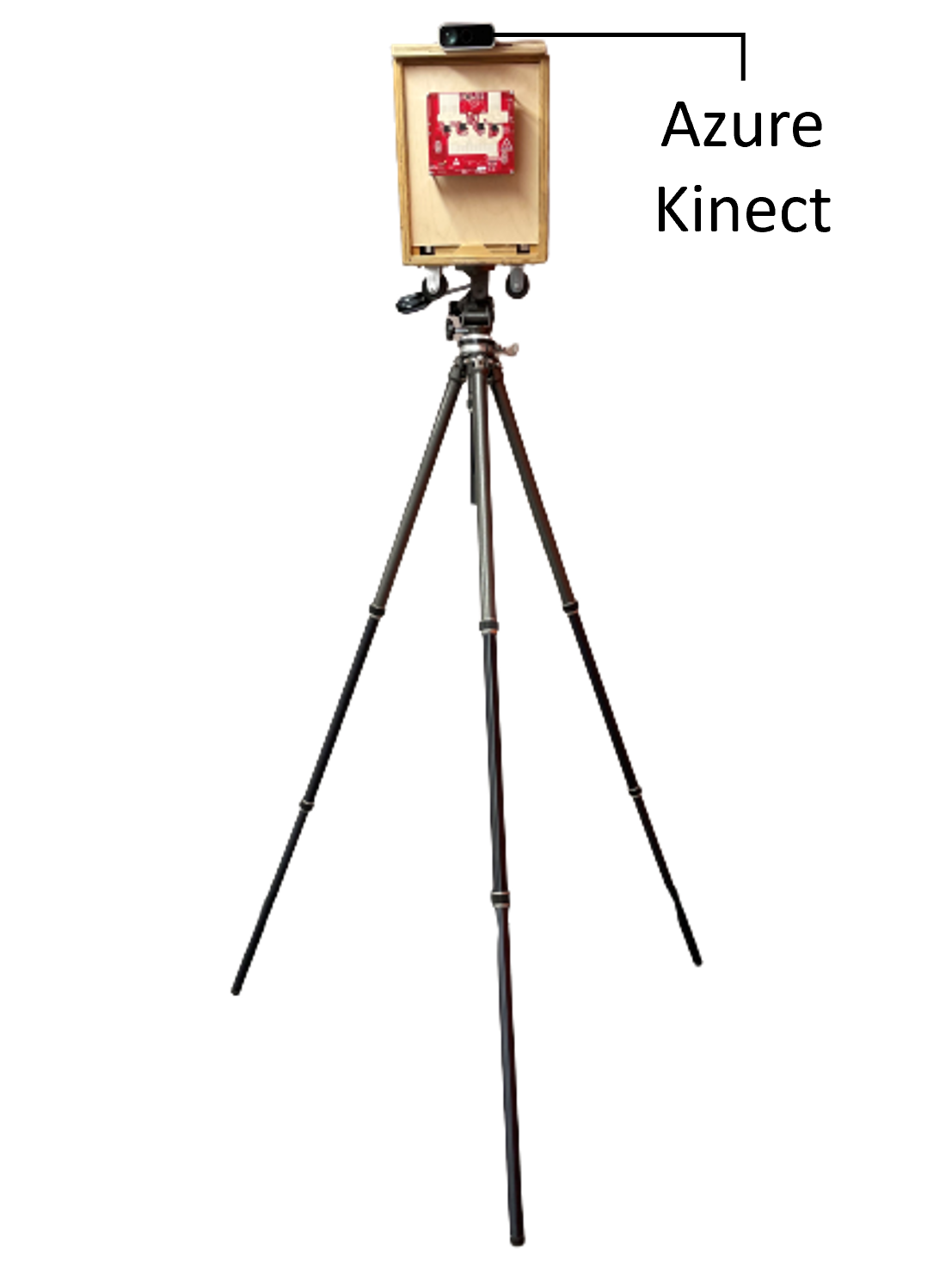} & 
             \includegraphics[height=5cm, width=3.5cm]{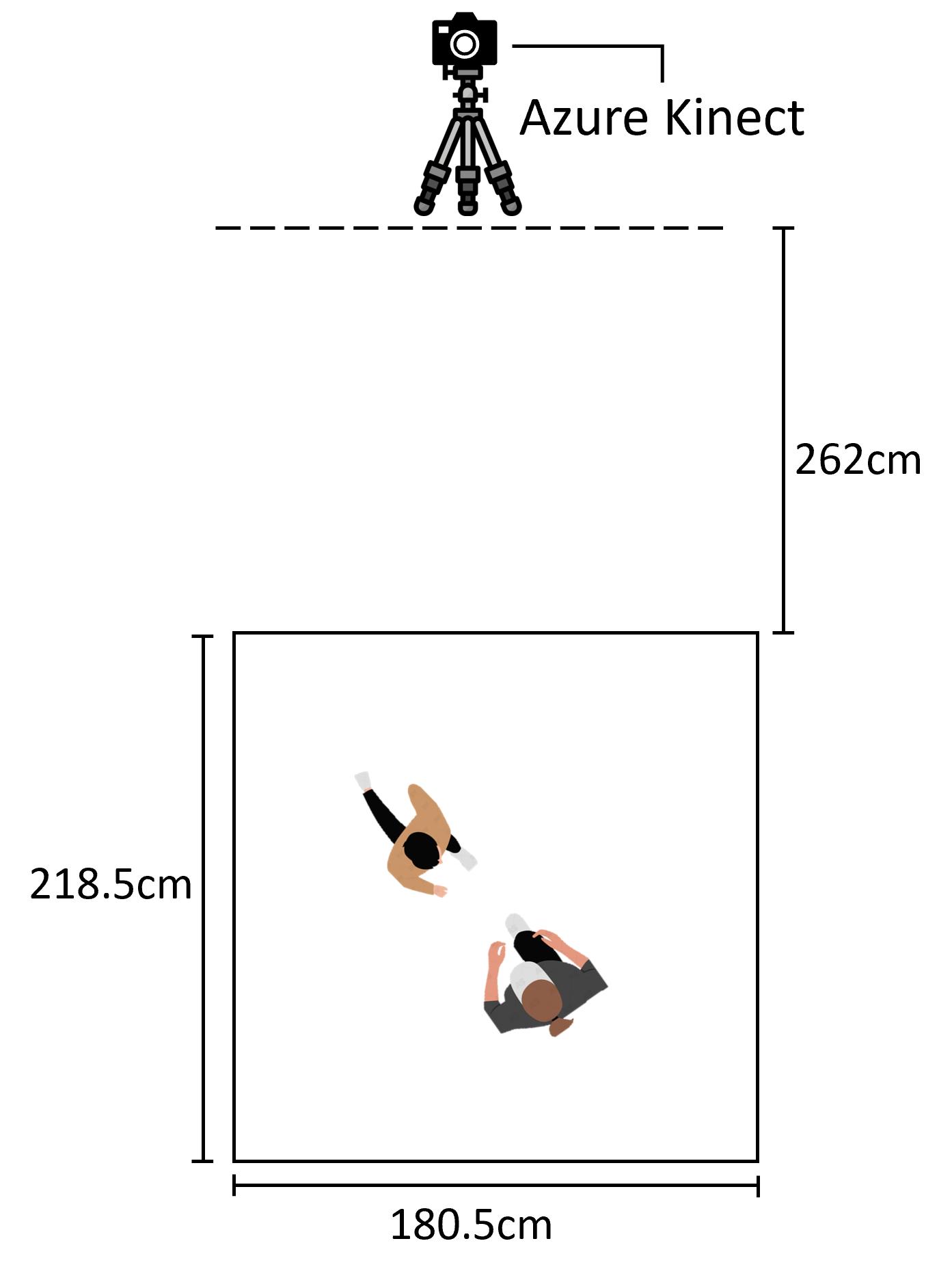}  &
             \includegraphics[height=5cm, width=3.5cm]{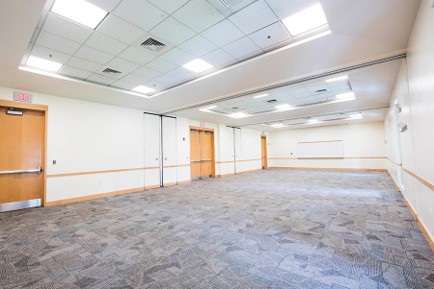}       \\
             (a) & (b) & (c)
        \end{tabular}
    \end{center}
    \vspace{-0.6cm} \caption{\justifying \textbf{The testbed includes (a) the Azure Kinect 221cm above ground and tilted 37 degrees downward, (b) a 218.5cm \(\times\) 180.5cm interaction area that is set 262cm away from the sensing module to ensure full view, in (c) an open indoor testbed location.}}
    \label{fig:testbed_setup}
\end{figure}

\vspace{-0.3cm} \begin{itemize}
    \item Emblems: Emblems are gestures that have a direct verbal translation and are usually culturally specific. The same gesture might be interpreted differently for different cultures \cite{hartman_nonverbal_nodate}. 
    Interactions chosen are \textbf{waving in}, \textbf{thumbs-up}, and \textbf{hand waving}.
    
    \item Illustrators: Bodily movements that illustrate the verbal message they accompany are called illustrators. They are used to clarify the conversation and are context dependent, meaning they lose their definitions if used independently \cite{chute_42_2023}. Interactions chosen are \textbf{pointing} and \textbf{showing measurements}.
    
    \item Affect displays: Affect displays are gestures that reveal one's affective and emotional state. 
    Interactions chosen are \textbf{hugging}, \textbf{laughing}, and \textbf{arm crossing}.
    
    \item Regulators: During interactions, regulators determine the alternation of instigating and receiving. Interactions chosen are \textbf{nodding} (i.e., acceptance or acknowledgement), \textbf{writing circles in the air} (i.e., sign to expedite), and \textbf{holding palms out} (i.e., sign to cease). 
    
    \item Adaptors: Adaptors are habitual movements that satisfy personal needs and can be used to increase or decrease emotional stability \cite{neff_dont_2011}. Interactions chosen are \textbf{twirling or scratching hair} to moderate one's stress during contemplation.
\end{itemize}

\section{Data Acquisition}
\vspace{\baselineskip}
\noindent \textbf{Data Collection Setup.} In this experiment, 4800 three-second samples of depth data were collected. These samples were carried out by 10 pairs of volunteers (i.e., 20 people in total). Every pair of participants performed 12 interactions for 40 repetitions each, all of which were performed at different angles with respect to the depth camera. All samples were taken in an open indoor space at CMU using a Microsoft Azure Kinect DK depth camera with frame rate 30 fps, resulting in 91 frames per sample. The testbed configuration is illustrated in Figure \ref{fig:testbed_setup}.
\vspace{4pt}
\section{Data Processing}
\vspace{\baselineskip}
\noindent \textbf{Feature Extraction.} Azure Kinect Body Tracking SDK is used to calculate two skeleton representations of participants from the depth data. Skeletal data of each sample are arranged in a \(91\times467\) array, where each row represents one frame and each column represents a feature extracted from the two skeletons. Features in each frame are organized as follows:
\begin{align*}
    [&J_0^1(x_t,y_t,z_t)..., V_0^1(x_t,y_t,z_t)...,\theta_{0t}^1,\omega_{0t}^1,C_{0t}^1...,D_{0t}^1...,\\
    &J_0^2(x_t,y_t,z_t)..., V_0^2(x_t,y_t,z_t)...,\theta_{0t}^2,\omega_{0t}^2,C_{0t}^2...,D_{0t}^2..., D_{bt}]
\end{align*}

\begin{figure} [t]
    \begin{center}
        \begin{tabular}{c c c}
             \includegraphics[height=5cm, width=5cm]{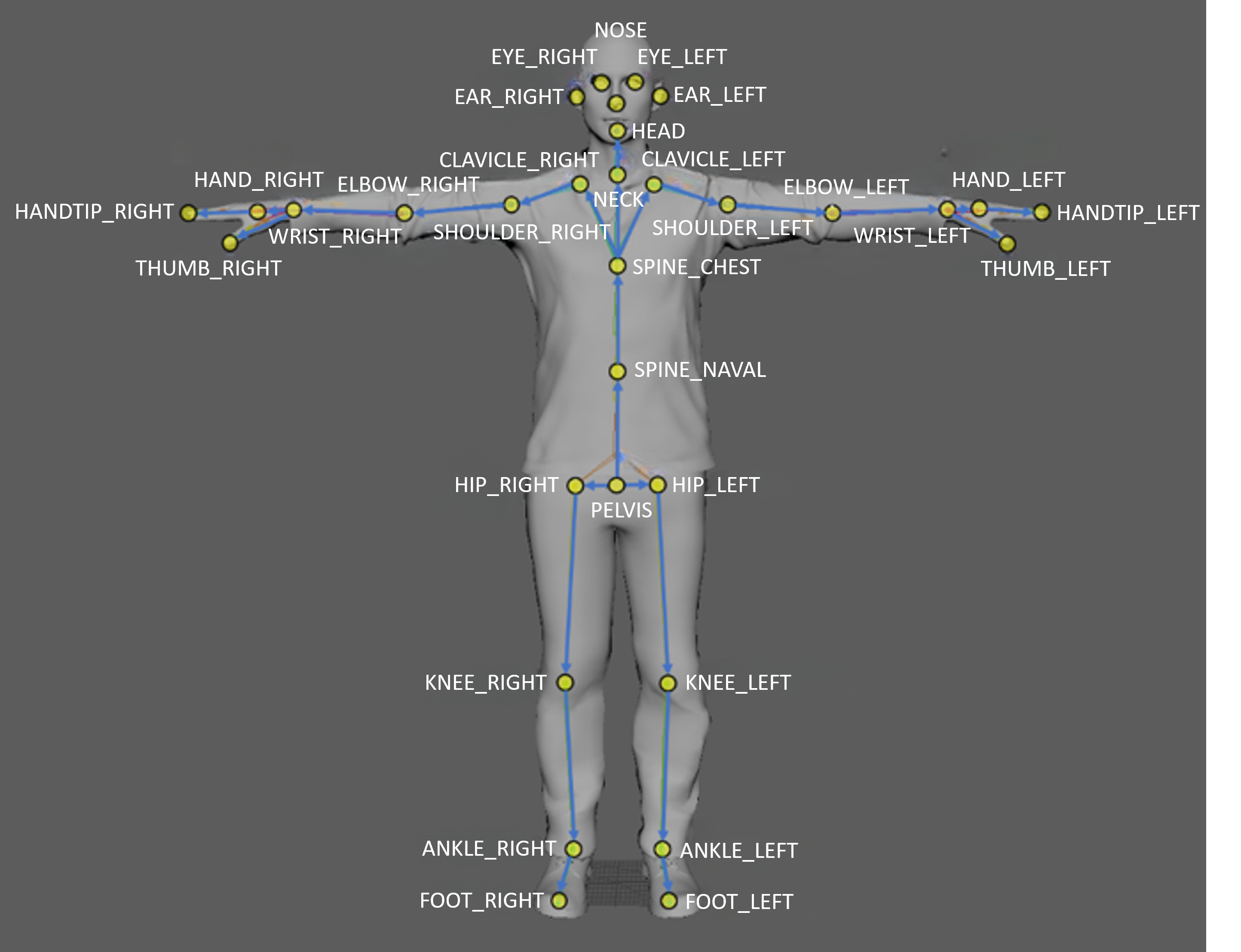} &
             \includegraphics[height=5cm, width=5cm]{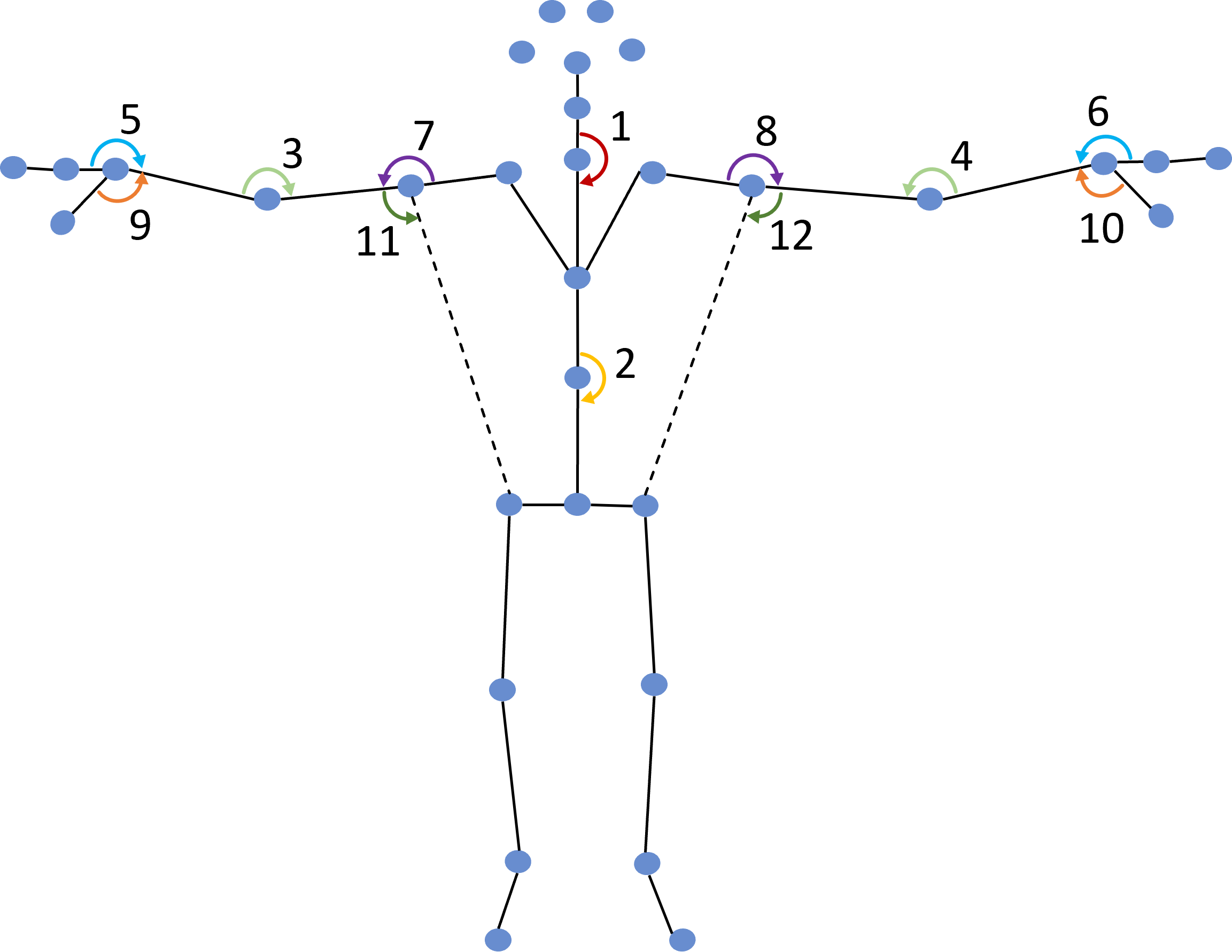}        &
             \includegraphics[height=5cm, width=5cm]{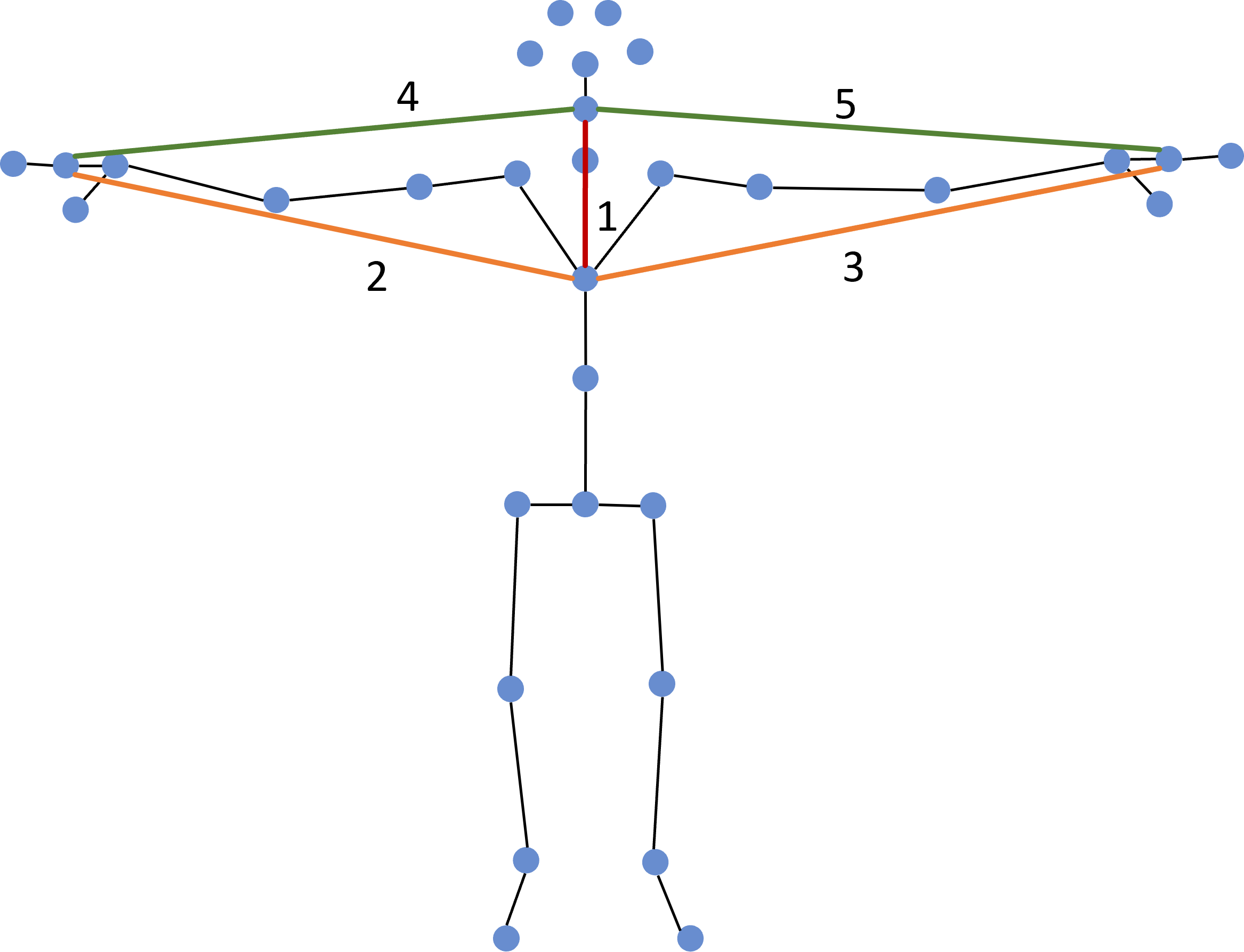}         \\
             (a) & (b) & (c)
        \end{tabular}
    \end{center}
    \vspace{-0.5cm} \caption{\textbf{The features extracted from skeletons are (a) 32 joints, (b) angles defined by three joints, and (c) distances that connect two joints.}\vspace{-0.5cm}}
    \label{fig:skeleton}
\end{figure}

\begin{itemize}
    \item Joint coordinates: The Azure Kinect SDK generates the 3D coordinates of 32 joints for each skeleton (Figure \ref{fig:skeleton}a). The origin of these coordinates is located at the focal point of the camera, which can induce significant noise resulting from the constant movement of participants. To mitigate noise, we translate the origin of each skeleton to their respective SPINE\(\_\)NAVAL joints \cite{ye_deep_2021}. Joint coordinates are denoted as \(J_n^m(x_t,y_t,z_t)\), where \(n\) is the joint number, \(m\) is the participant, and \(t\) is the frame number (the use of \(m\) and \(t\) persists through the section). Overall, we acquire 32 3D coordinates of two skeletons, totaling 192 features from joint coordinates in one frame. 
    \item Velocities: Velocity is defined as the rate of motion of each joint, which is determined by the displacement of each joint divided by the time elapsed between two consecutive frames. \(V_n^m(x_t,y_t,z_t)\) represents the velocity of the \(n^{th}\) joint of the \(m^{th}\) participant between frames at \(t-1\) and \(t\). Each frame provides 192 velocity-related features, which are composed of the velocities of 32 joints of two participants in 3D space. 
    \item Angles, angular velocities, and confidence level: 12 angles (Figure \ref{fig:skeleton}b) are included in the extracted features because bending motions are present in the interactions chosen (especially in the upper body). Angles are subtended by two vectors joining at a joint, each of which is extended from the joined joint to an adjacent joint. For example, Angle 3 is subtended by two vectors joining at ELBOW\(\_\)RIGHT, one of which is extended from ELBOW\(\_\)RIGHT to WRIST\(\_\)RIGHT and the other is extended from ELBOW\(\_\)RIGHT to SHOULDER\(\_\)RIGHT. Angular velocities are obtained by dividing the difference of the angle between previous and current frames by the time passed. These two features, along with confidence level of the joined joint, are extracted as a set. \(\theta^m_{kt}\), \(\omega^m_{kt}\), and \(C^m_{kt}\) are angle, angular velocity, and confidence level respectively, where \(k\) indicates the angle number. In total, we extract 24 angles, 24 angular velocities, and 24 confidence levels from both participants in each frame.
    \item Distances: In each frame, we define 5 intra-body distances for each skeleton (Figure \ref{fig:skeleton}c) and one inter-body distance connecting the SPINE\(\_\)NAVAL joints of the two skeletons, providing 11 features per frame. In terms of notation, \(D^m_{jt}\) and \(D_{bt}\) denote the \(j^{th}\) chosen distance of the \(m^{th}\) participant in frame \(t\) and the distance between two skeletons respectively. 
\end{itemize}

\begin{figure} [t]
    \begin{center}
        \includegraphics[scale = 0.47]{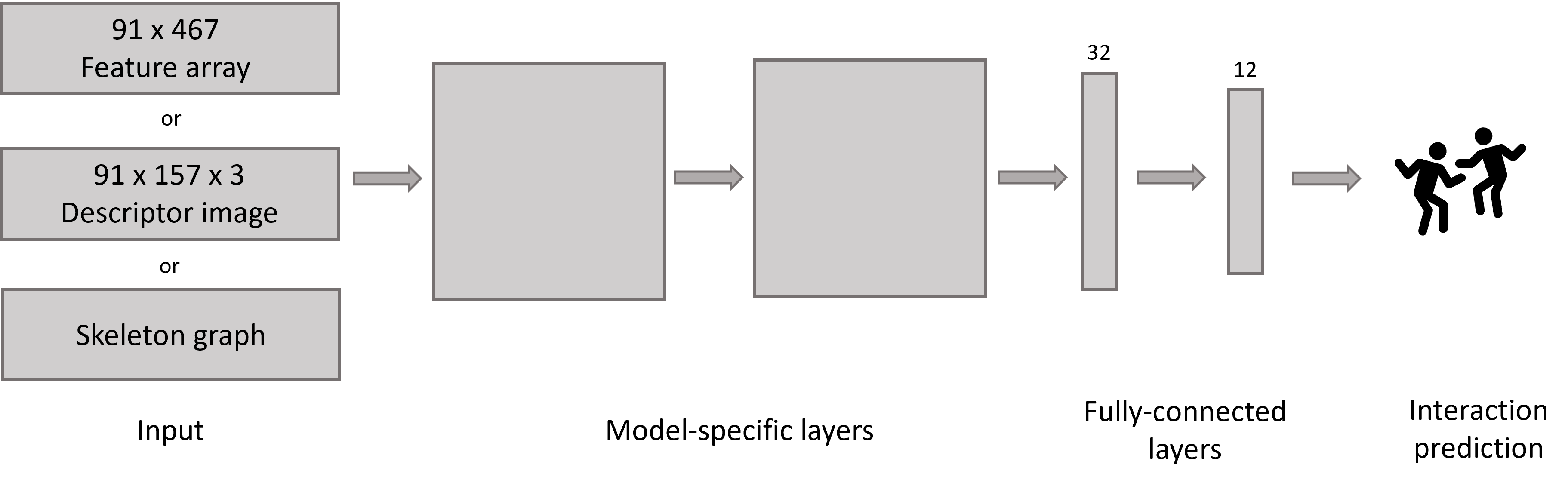}
    \end{center}
    \vspace{-0.5cm} \caption{\textbf{Overall structure of the deep learning models, which includes the input layer (depends on the considered model), model-specific layers, two fully connected layers containing 32 and 12 neurons respectively, and the output layer with softmax activation function.}\vspace{-0.5cm}}
    \label{fig:deep_learning_structure}
\end{figure}

\noindent \textbf{Skeleton Descriptor Images.} After extracting the desired features from the skeletons, the features are translated to different input forms depending on the deep learning algorithm considered, as demonstrated in Figure \ref{fig:deep_learning_structure}. For instance, the aforementioned array can be used as input for Bi-dir LSTM, ConvLSTM, transformer directly. CNN models require descriptor images. One descriptor image represents one video sample, which is shaped 91\(\times\)157\(\times\)3. Each row of a descriptor image delineates a frame, each column represents features related to one of the joints, and the 3 channels are $x$, $y$, and $z$ coordinates of the joint or a set of three features. For instance, the $x$, $y$, and $z$ coordinates of SPINE\(\_\)NAVAL in the first frame are mapped to the first, second, and third channels in the first row and second column. The first 32 columns are the 3D coordinates of 32 joints of the first subject. The second 32 columns capture velocities of 32 joints in 3D representation of the first subject. Following are 12 columns of angle-related features of 12 joints, where the three channels are angle, angular velocity, and the confidence level of the chosen joint, respectively. 5 intra-distances of the first subject constitute the next 2 columns, where the missing channel is replaced by a zero. Features of the second subject follow the previous arrangement and produce the next 78 columns. The last column in a row is the inter-distance between the two skeletons and two zeros to make up for the missing two channels.

\vspace{\baselineskip}
\noindent \textbf{Graphical Representation.} GCN-based models (e.g., ST-GCN) require graphs as inputs, which are composed of nodes and edges. Nodes are embedded with feature vectors and edges are connections between two nodes. The graphical representation of skeletal features encode joints as nodes, which vectorize and store features (e.g., velocity and angle) of the corresponding joint in each node. There are three embeddings for edges: natural, temporal, and inter-body connections. The distances of natural connections between joints, as shown in Figure \ref{fig:skeleton}a, are represented by edges, which account for the spatial correlation between joints during an interaction. For temporal correlation, we link the same joint in adjacent frames with edges to observe the displacement of joints in different frames. Another edge that connects the SPINE\(\_\)NAVAL of two skeletons represents the interactive element of interactions, which is unique to dyadic interactions.

\vspace{4pt}
\section{Results and Discussion}
\begin{figure} [b]
\vspace{-0.3cm}
    \begin{center}
        \includegraphics[scale = 0.4]{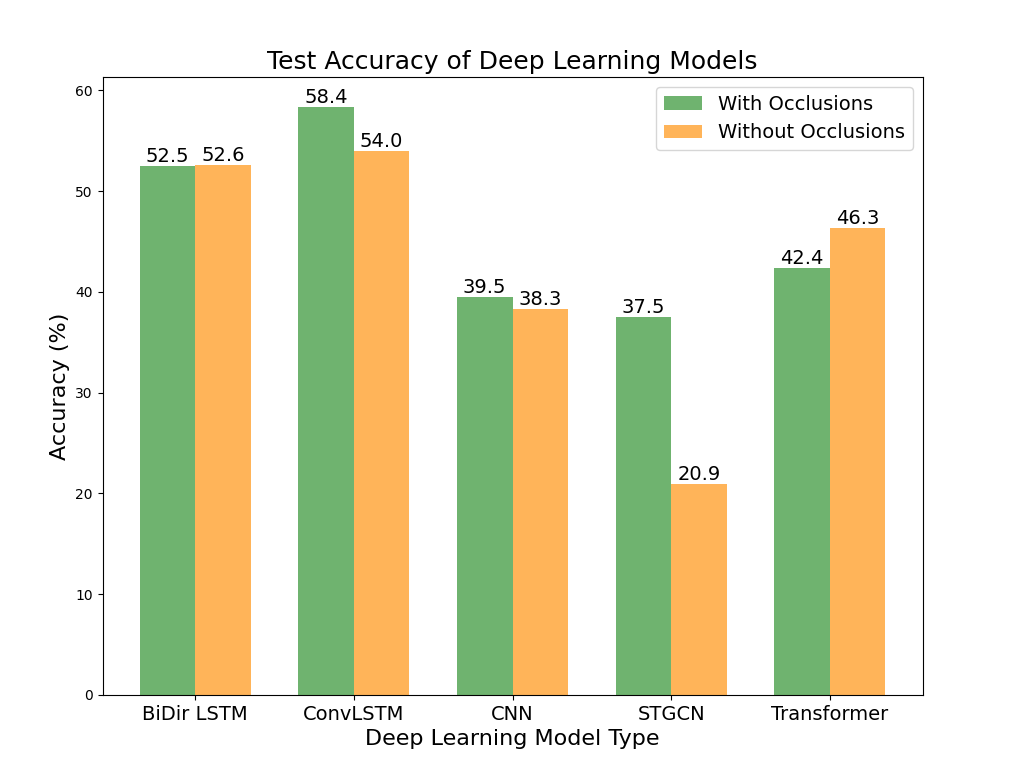}
    \end{center}
    \vspace{-0.5cm} \caption{\textbf{Test accuracy of the deep learning models. Both tests with and without occlusions are considered. Here, occlusions occur when one participant occludes the other participant.}\vspace{-0.5cm}}
    \label{fig:results}
    \vspace{-0.3cm}
\end{figure}

\vspace{\baselineskip}
\noindent This work benchmarks five deep learning algorithms for HIR to justify the choice of a specific model for dyadic interaction recognition, as well as to provide insights on future research directions. Overall, ConvLSTM outperforms all other models (Figure \ref{fig:results}), as it combines the strengths of LSTM and CNN and embraces the spatiotemporal nature of interactions. Additionally, LSTM-based models require little data processing after feature extraction (compared to converting to descriptor images and graphical representations), which is computationally efficient in real-life cases. Another model that combines the temporal and spatial strong suits is ST-GCN. However, our results indicate that it has the lowest accuracy. This is justified by the model's difficulty in observing the displacement of a specific joint across all frames due to the graphical representation, which only connects the same joint in consecutive frames. This temporal linkage contrasts with LSTM-based models, which better tracks joint movement over time. The importance of temporal correlation is accentuated by the comparison between Bi-dir LSTM and CNN. Bi-dir LSTM performs better than CNN even though the ratio of temporal to spatial features in our data is approximately 1 to 5. Transformer is the third-best performing model in this work, but its computationally-expensive characteristic undermines its applicability for real-world applications. 

Through this benchmarking we also gain insight into the influence of occlusions on model performance. As dyadic interactions require monitoring the behaviors between two or more people, occlusions have greater implications here than in the monadic case. Figure \ref{fig:results} indicates that occlusions do not heavily affect the skeleton-based interaction recognition models, which is a common issue with HIR. Occlusion happens when one subject is obscured by the other, leaving only one subject visible within the sensor's view. We explore the effect of occlusions on different HIR models by training 5 algorithms with two batches of samples separately: one batch includes a mixture of both occluded and unoccluded videos, and the other batch contains only unoccluded videos. In frames where occlusion occurs, all features of both subjects in those frames are assigned with zeros since partial information of an interaction can be misleading. From the results shown in Figure \ref{fig:results}, no prominent accuracy disparity is found between occluded and unoccluded data, except for ST-GCN. ST-GCN has better performance for the occluded data because zeros representing occlusions desensitize the model from small changes originating from the individuality of subjects \cite{ali_skeleton-based_2023}. This finding can help to inform testbed configurations  (i.e., the positioning of depth sensors). Unlike the complex setups required with multiple cameras around a test area to avoid occlusions for RGB-based methods, skeleton-based HIR for monitoring dyadic interactions simplifies the process and, in part, overcomes occlusion issues.

Monadic actions can also leverage these 5 considered algorithms. \citeN{martins_skeleton-based_2023} compared the performances of these models on monadic actions pertaining to thermal comfort, which produces higher accuracy than our work. The difference in accuracy can be induced by the sheer complexity of dyadic interactions. Measuring dyadic interactions must capture communication between two individuals through bodily movements, whereas monadic actions involve the actions of only one subject. The inclusion of one or more additional people produces another degree of uncertainty introduced by the individuality of each subject to the recognition task. While each monadic action has only a few patterns for each input, every dyadic interaction contains an assortment of patterns for one input, which is another factor contributing to the disparity in accuracy between monadic and dyadic interaction recognition. Additionally, a dyadic interaction is formulated by an instigator (i.e., the subject initiating the communication) and a receiver (i.e., the subject reacting to the communication). In our work, the input data that records the social exchange can be preceded with the instigator or the receiver. For instance, the first 233 columns of a feature array can either describe features of an instigator or a receiver, and the rest for the other. This generates multiple patterns for a single interaction label, which subverts the generalizability of the deep learning models, but adheres to real-world scenarios of dyadic interaction recognition where people most likely switch between an interaction instigator and receiver multiple times between actions. As a key takeaway, developing powerful models for understanding dyadic interactions is essential to garnering higher-level insights into human behaviors and social impacts tied to CPSIS.

\section{Conclusion}
\vspace{\baselineskip}
\noindent While the incorporation of sensing and control has enhanced infrastructure performance, there remains an uncharted area with the potential to enhance the social benefits of infrastructure systems (termed CPSIS). To advance the study of CPSIS, there is a need to identify and measure human-human and human-infrastructure interactions accurately and in a privacy-preserving way. Rather than attempting to exhaust all human interactions, we develop a dataset containing 12 interactions from a taxonomy of interactions based on their communication functions. We then benchmark 5 widely-used deep learning models to justify the use of specific models for dyadic interaction monitoring. ConvLSTM emerges as the most suitable method for extracting social interactions.

Next steps of this work can be continued in two primary directions: identify the distinctive characteristic of social interactions and compiling the embedded meanings of our dataset. The 5 selected deep learning models have been benchmarked on popular datasets (e.g., NTU RGB+D \cite{shahroudy_ntu_2016}) separately and obtained decent accuracy. However, the disparity in accuracy between our dataset and those well-received monadic ones illustrates that social dyadic interactions possess a unique characteristic that has not been addressed by existing work. This reinforces the need to identify the nuanced quality of social interactions and ameliorate these deep learning algorithms through more intricate data extraction and post-processing strategies to study these interactions more accurately and increase the accuracy of the recognition task. Deeper comprehension at this interaction level would enable the establishment of a framework capable of mapping social interactions to social objectives. Such a mapping could be leveraged by a multitude of human-centered applications where social objectives and privacy weigh highly. Applications that would benefit from such a mapping include, but are not limited to, healthcare, autonomous vehicles, smart homes, and social infrastructure systems.
\vspace{7pt}
\section{Acknowledgements}

\noindent The authors gratefully acknowledge Brian Belowich and Sirajum Munir for their assistance in configuring and providing insights on the testbed setup.  This work is funded by the Pennsylvania Infrastructure Technology Alliance, CMU Manufacturing Futures Institute, and the Mobility21 United States Department of Transportation University Transportation Center.

\bibliography{References}

\end{document}